\documentclass[runningheads]{llncs}

\usepackage{eccv}

\usepackage{eccvabbrv}

\usepackage{graphicx} 
\usepackage{wrapfig}     
\usepackage{booktabs} 
\usepackage{amsmath}  
\usepackage{amsfonts} 
\usepackage{amssymb}  
\usepackage{multirow} 
\usepackage{tabularx} 
\usepackage[table,xcdraw]{xcolor} 
\usepackage[accsupp]{axessibility}  

\usepackage{orcidlink}
\usepackage{pifont}

\definecolor{HeaderGray}{RGB}{235, 235, 240} 

\definecolor{LightBlueRow}{RGB}{248, 252, 255} 

\definecolor{OursHighlight}{RGB}{245, 240, 255} 

\definecolor{DeepBlueText}{RGB}{0, 70, 140} 

\usepackage{hyperref}

\def\ourmodel{VideoTIR }
\newcommand{\equalcontrib}{\textsuperscript{*}}
\newcommand{\corrauthor}{\textsuperscript{\ding{41}}}
\newcommand{\projectlead}{\textsuperscript{\textdagger}}

\begin{document}
\title{VideoTIR: Accurate Understanding for Long Videos with Efficient Tool-Integrated Reasoning} 

\titlerunning{VideoTIR}

\author{Zhe Gao\inst{1}\equalcontrib \and
Shiyu Shen\inst{2}\equalcontrib \and
Taifeng Chai\inst{3} \and
Weinong Wang\inst{4}\projectlead \and
Haotian Xu\inst{5} \and
Xing Wu\inst{4} \and
Wenbin Li\inst{1} \and
Qi Fan\inst{1}\corrauthor \and
Yang Gao\inst{1} \and
Dacheng Tao\inst{6}}

\authorrunning{Z.~Gao et al.}

\institute{Nanjing University, Nanjing, China\\
\email{fanqi@nju.edu.cn} \and
Nankai University, Tianjin, China \and
East China Normal University, Shanghai, China \and
Tencent Hunyuan, China \and
MiroMind, USA \and
Nanyang Technological University, Singapore}

\maketitle
\begingroup
\renewcommand{\thefootnote}{}
\footnotetext{\equalcontrib Equal contribution. \corrauthor Corresponding author. \projectlead Project leader.}
\endgroup
\begin{abstract}
Existing Multimodal Large Language Models (MLLMs) often suffer from hallucinations in long video understanding (LVU), primarily due to the imbalance between textual and visual tokens.
Observing that MLLMs handle shorter but more accurate visual inputs well, recent LVU works alleviate hallucinations by automatically parsing the vast visual data into manageable segments that can be effectively processed by MLLMs.
SFT-based tool-calling methods can serve this purpose, but they typically require vast amounts of fine-grained, high-quality data and suffer from constrained tool-calling trajectories.
We propose a novel \ourmodel that leverages Reinforcement Learning (RL) to encourage proper usage of comprehensive multi-level toolkits for efficient long video understanding.
\ourmodel explores both Zero-RL and SFT cold-starting to enable MLLMs to retrieve and focus on meaningful video segments/images/regions, enhancing long video understanding both accurately and efficiently.
To reduce redundant tool-calling in the early RL-stage and accelerate convergence, we propose Toolkit Action Grouped Policy Optimization (TAGPO), which enhances the efficiency of the calling process through the finer stepwise reward assignment.
Additionally, we develop a sandbox-based trajectory synthesis framework to generate high-quality trajectory data.
Extensive experiments on three long-video QA benchmarks demonstrate the effectiveness and efficiency of our method.


\keywords{Long Video Understanding \and Vision-Language Models \and Agentic RL \and Tool-Integrated Reasoning}
\end{abstract}

\section{Introduction\label{sec:intro}}

\begin{figure}[ht]
\centering
\includegraphics[width=\columnwidth]{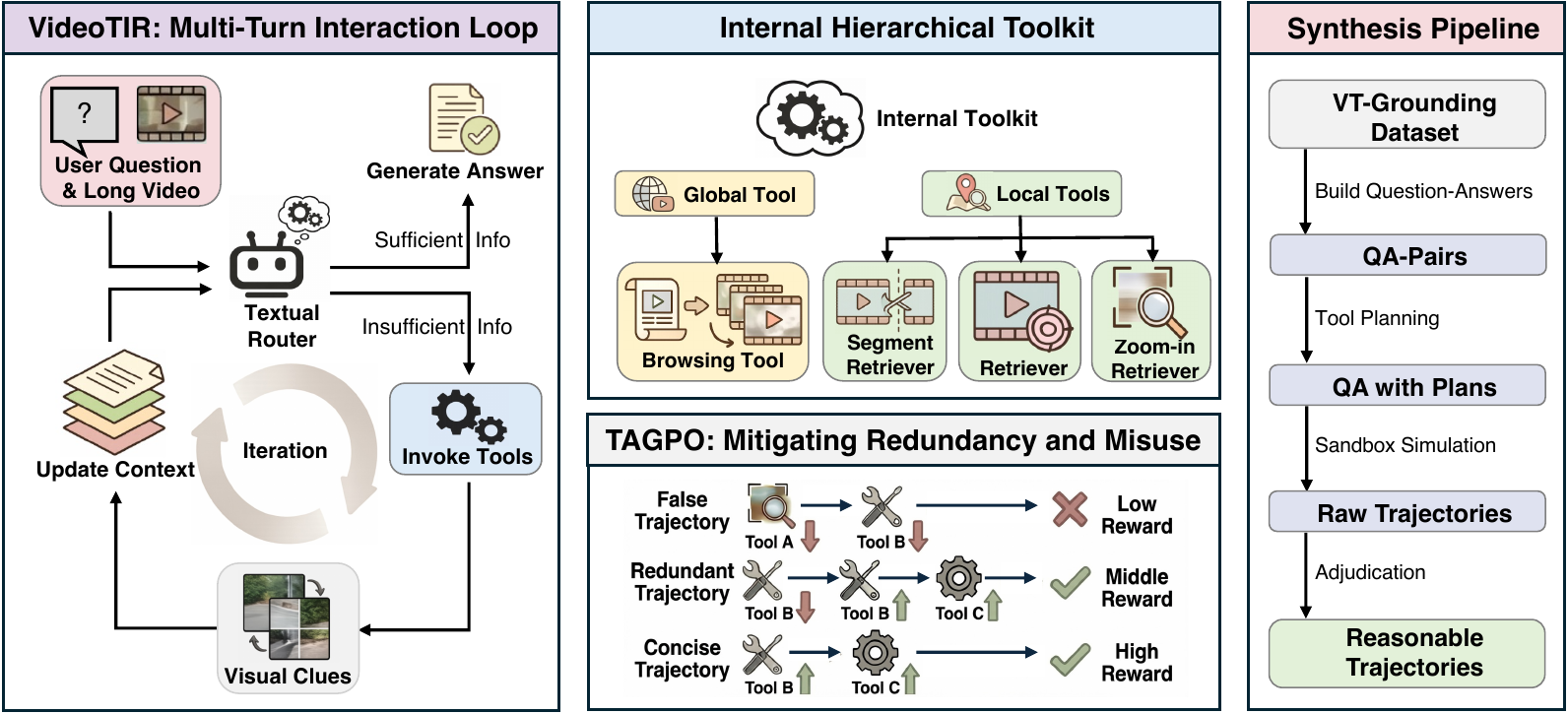}
\caption{We propose \ourmodel, a tool-integrated reasoning framework that flexibly and hierarchically retrieves relevant video segments through endogenous tool invocation to support long-video understanding. Furthermore, to enable SFT cold start, we introduce a sandbox-based trajectory synthesis framework. We also present TAGPO to address the inefficiency in early-stage RL exploration caused by tool misuse and overuse.}
\label{fig:teasure}
\end{figure}

Long video understanding (LVU) \cite{wang2025lvbench,wu2019long, wu2021towards} aims to answer user questions based on long videos, whose durations typically span several minutes to multiple hours.
To tackle this challenge, this field has increasingly turned to Multimodal Large Language Models (MLLMs \cite{weng2024longvlm,qwen25,wang2024qwen2,glm,internvl}), leveraging their powerful vision-language reasoning capabilities to interpret complex, long-form video content.

Recently, long video understanding methods can be broadly categorized into non-TIR frame selection strategies and tool-integrated reasoning (TIR) approaches.
Non-TIR methods, such as frame selection strategies \cite{hu2025m}, select relevant frames before MLLM final reasoning.
While effective in reducing redundant visual tokens, these approaches operate outside the reasoning loop and lack adaptive refinement during inference.
In contrast, TIR methods integrate retrieval tools into the reasoning process, enabling iterative evidence acquisition.
Existing TIR designs generally fall into two categories.
The first category relies on heavy external retrieval toolkits \cite{videomind, videoagent, fan2024videoagent, ma2025drvideo}, which often involve complex pipelines and fixed action workflows.
Such rule-based or externally dependent systems suffer from limited generalization and high interaction overhead.
The second category designs lightweight internal retrieval mechanisms by prompting MLLMs to output textual timestamps that guide segment extraction \cite{videomtr, longvt}.
However, as illustrated in~\cite{molmo}, current MLLMs are not sufficiently pre-trained on video datasets with fine-grained spatiotemporal annotations. Consequently, both zero-RL strategies~\cite{videomtr} and RL-after-SFT cold-start designs~\cite{longvt} struggle to substantially improve localization precision due to the limited capabilities of the base model.
As a result, retrieval often becomes redundant or inefficient, with repeated tool invocation over similar segments.
Moreover, these methods rely on a single-tool toolkit, which limits their ability to flexibly handle video question answering tasks.

Therefore, we aim to design a diverse and efficient internal toolkit that leverages the MLLM’s own encoders. In addition, we propose an RL framework to effectively coordinate these tools, enabling flexible and question-aware tool invocation.
Detailedly, we propose \ourmodel, a novel approach for long video understanding using Tool-Integrated Reinforcement Learning.
VideoTIR comprises two key components: a comprehensive toolkit containing multiple specialized tools for diverse sub-tasks and a textual router that dynamically selects a subset of candidate tools from the toolkit.
The toolkit includes diverse specialized tools for both global and local video processing tasks.
For global tasks, we introduce a browsing tool that progressively increases video spatial resolution and sampling rate to acquire coarse-to-fine visual evidence accordingly.
For localized queries, we introduce a fine-grained retrieval tool chain comprising clip-segment, frame-pick, and zoom-in tools, enabling multi-granularity and flexible retrieval to support temporally and spatially specific tool-calling requirements.
This multi-level toolkit provides flexible support for adaptive decision-making across heterogeneous questions in long videos.
Leveraging the extensive toolkit, the textual router utilizes the reasoning capabilities of MLLMs to understand long videos by pre-scanning them, interpreting questions, and formulating subsequent action.

However, we observe that single-tool Tool-Integrated RL methods~\cite{deepeyes,thyme,videomtr}, which typically adopt GRPO as the policy update strategy and design rewards primarily around final answer accuracy and tool invocation, often lead to both \textbf{misuse} and \textbf{overuse} of tools within the multi-tool settings.
Misuse refers to invoking multiple tools while still producing incorrect answers. This behavior reflects reward hacking and increases the trajectory exploration burden during the early stages of RL training, resulting in excessive tool invocation.
In contrast, overuse in our context refers to the tendency of MLLMs to retrieve the finest-grained frames within the retrieval tool chain, even when coarser-grained retrieval would be sufficient to answer the query.
To address these issues, we propose Toolkit Action Grouped Policy Optimization (TAGPO), a novel RL optimization algorithm that improves the efficiency of tool utilization through stepwise reward assignment.
Specifically, TAGPO groups invocations of the same retrieval sub-tool and designs an advantage estimation scheme that assigns higher importance to sub-tool invocations closer to the final answer, thereby mitigating tool overuse.
Moreover, this advantage formulation assigns lower advantage to trajectories that reuse prior experience—i.e., invocation chains that led to correct answers in other rollouts—but still produce incorrect outputs, compared to trajectories that explore new tools, thereby reducing tool misuse.

Moreover, multi-tool invocation introduces new challenges for accurate instruction following in multi-tool agents.
Multi-tool reasoning typically relies on heterogeneous interactions across various tools, where tool prompts may be invoked far from the initial system prompt, increasing the risk of instruction drift or forgetting.
Although supervised fine-tuning (SFT) may alleviate this issue, it typically requires extensive high-quality multi-tool video datasets, which are often difficult to obtain.
To address this problem, we propose a data synthesis framework to bootstrap multi-instruction-following capabilities for RL agents.
We leverage an external MLLM (\textit{e.g.}, GLM‑4.5V~\cite{glm}) to generate question–answer pairs and explore possible tool invocation orders based on a video–text grounding dataset.
We then construct a sandbox to enable MLLMs to generate additional synthetic training data, including intermediate prompts and synthesized trajectories.
The synthesized data can be used to pretrain instruction-following capabilities and tool-use policies for agents.

Our main contributions are summarized as follows (Fig.~\ref{fig:teasure}):
\begin{itemize}
    \item \textbf{Multi-turn multi-internal-tool agent for LVU.} We propose a multi-turn, multi-tool agent framework to effectively and efficiently understand long videos by leveraging internal tools.
    \item \textbf{Invocation-aware reinforcement learning.} We design a novel toolkit action grouped policy optimization to explicitly encourage concise and effective tool calls, balancing video exploration efficiency and reasoning accuracy.
    \item \textbf{Multi-tool trajectory synthesis.} We develop and open-source a sandbox-based trajectory synthesis framework that equips LVU models with instruction-following and tool-calling abilities before RL fine-tuning.
\end{itemize}

\section{Related Works}
\label{sec:formatting}

\noindent\textbf{Multimodal Models in Video Understanding.}
Video understanding encompasses tasks such as action recognition, temporal localization and retrieval, captioning, and summarization. Early deep learning approaches typically targeted single tasks, with architectures designed to model temporal dependencies and extract spatiotemporal features \cite{wu2021towards, tong2022videomae}. With the advent of MLLMs, the field has shifted toward unified, task-agnostic frameworks. These systems commonly employ an LLM as the reasoning core while aligning video representations through a visual encoder and a multimodal connector \cite{alayrac2022flamingo}. Representative examples include Video-LLaMA \cite{zhang2023video}, which integrates image/video and audio branches with an LLM; and Video-ChatGPT \cite{maaz2024video} and Video-LLaVA \cite{lin2024video}, which build instruction-following conversational capabilities from curated video QA and caption datasets. More recent unified models \cite{wang2024qwen2} process images and videos using shared tokenization and temporal positional encodings, achieving competitive zero- and few-shot performance across diverse benchmarks. Comprehensive evaluations \cite{li2024mvbench, fu2025video} show that MLLMs perform strongly on short- to medium-length clips for perception and reasoning but degrade as clip duration increases, indicating a need for more robust temporal and multimodal integration.

Despite this progress, applying MLLMs directly to LVU exposes key limitations. First, constrained context windows necessitate aggressive subsampling, which obscures cross-scene dependencies and amplifies duration-induced errors \cite{mangalam2023egoschema}. Second, temporal reasoning and precise localization are fragile in the absence of explicit, retrievable memory, leading to ordering and causal mistakes \cite{xiao2021next}. Third, long-horizon datasets and annotations remain scarce and expensive, biasing instruction corpora toward short clips \cite{grauman2022ego4d}.

\begin{figure*}[!t] 
  \centering
  \includegraphics[width=\textwidth]{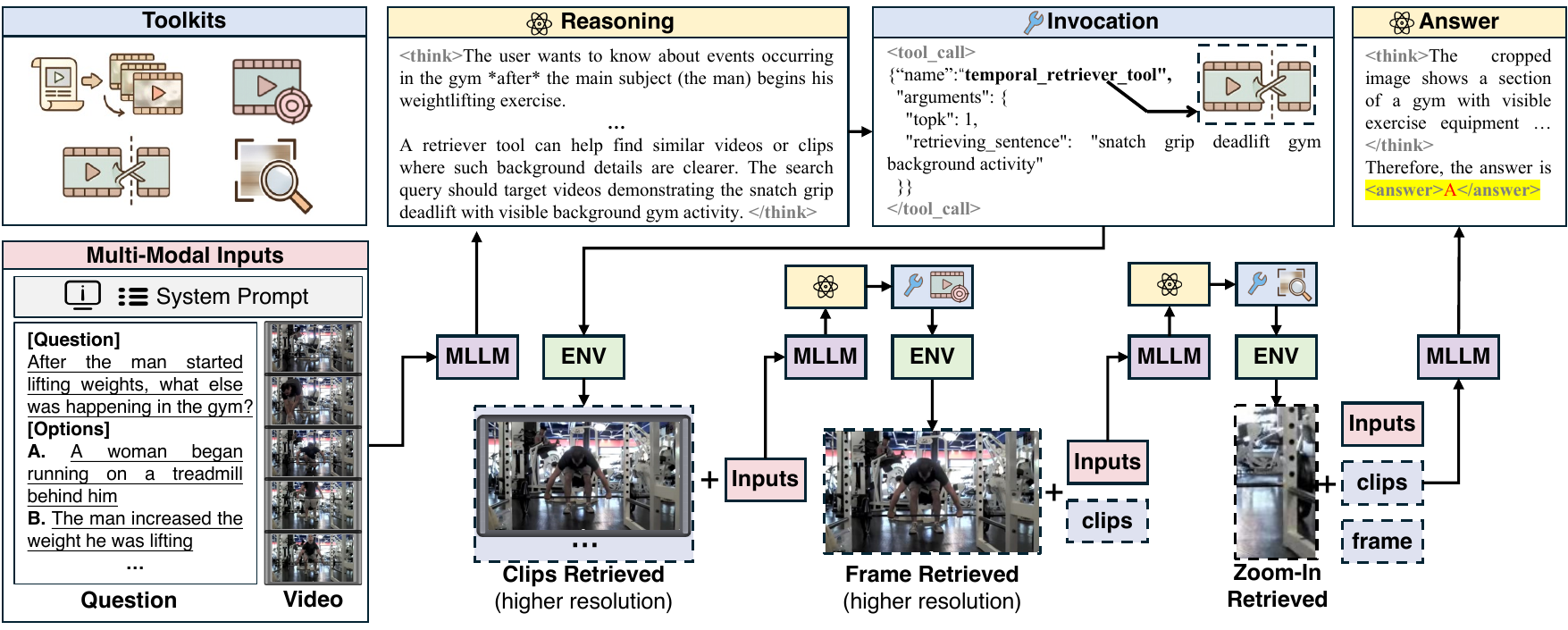}
  \caption{Framework of our methods. VideoTIR adopts a multi-turn manner to deal with the users' input videos and questions. When the model fails to conclude an answer based on current visual information, it calls tools to perceive the absent vision clues, which is combined with the former context as the input for the next-turn reasoning.}
  \label{fig:framework}
\end{figure*}

\begin{figure}[t]
\centering
\includegraphics[width=\columnwidth]{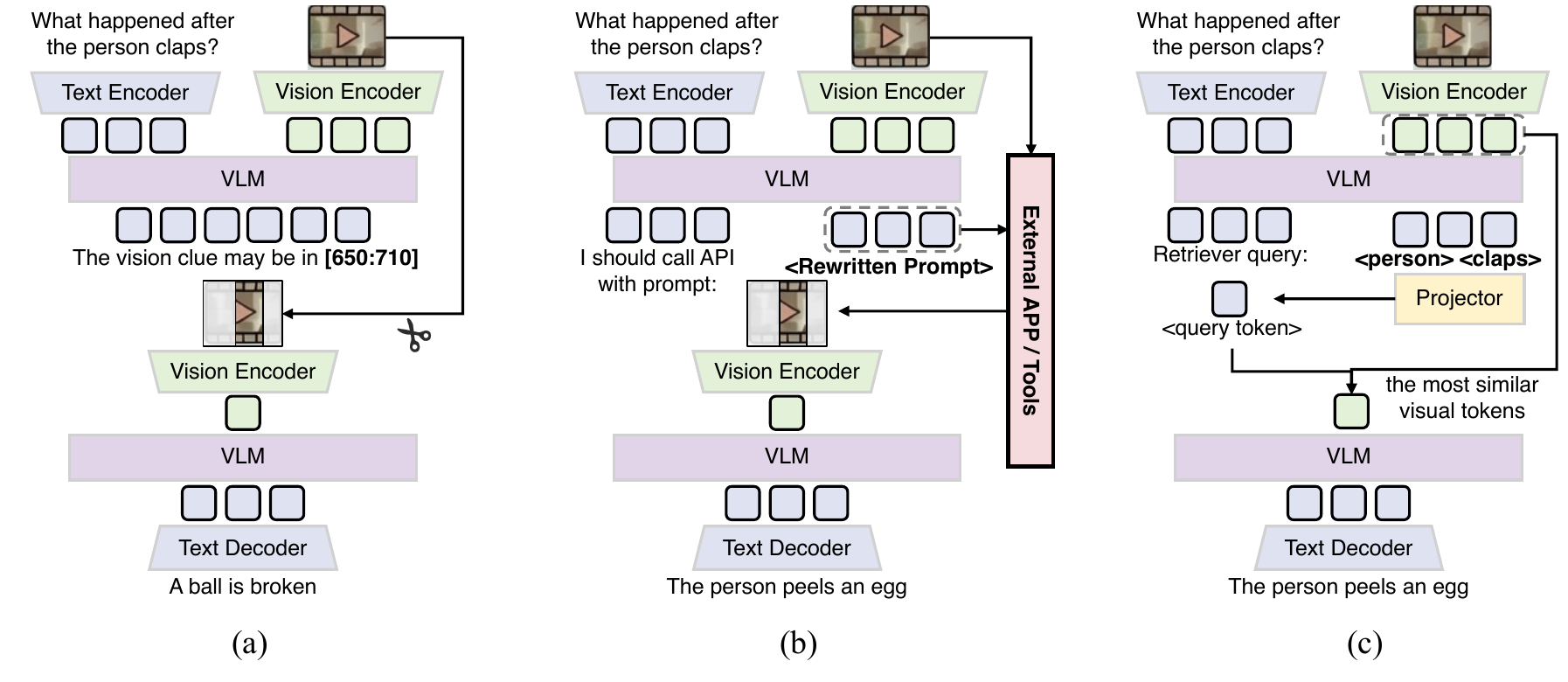}
\caption{Comparison of tool-integrated reasoning (TIR) designs for video understanding. (a) Methods such as \cite{longvt, videomtr} adopt a paradigm in which the VLM outputs timestamps in text form for subsequent video clipping. (b) Alternatively, some methods rely on heavyweight external tools, incurring substantial interaction costs. In contrast, \ourmodel leverages the intrinsic encoding structure of the VLM to design internal retrieval tools, selecting visual cues from the video based on feature similarity.}
\label{fig:tool_compares}
\end{figure}

\noindent\textbf{Tool-Based Models in Video Understanding.}
Recent progress in long video understanding can be organized along the spectrum from frame-selection strategies to fully tool-integrated reasoning (TIR) agents.
\textit{Frame-selection methods} select relevant frames or segments prior to final reasoning, operating largely outside the reasoning loop.
Such approaches reduce redundant visual tokens and improve efficiency, but lack iterative refinement and adaptive evidence acquisition.
\textit{Tool-integrated reasoning agents}, inspired by ReAct-style paradigms \cite{yao2022react, shinn2023reflexion}, instead interleave planning, tool invocation, and reasoning.
Within TIR approaches, existing systems can be broadly categorized into external-tool pipelines and lightweight internal retrieval designs.
External-tool agents couple MLLMs with structured retrieval modules such as ASR, OCR, segment localizers, or memory systems \cite{fan2024videoagent, ma2025drvideo, jeoung2024adaptive}.
These systems often rely on pre-defined workflows, multi-agent coordination \cite{chen2025lvagent, kugo2025videomultiagents}, or structured memory decomposition \cite{wang2025videochat, xu2025vrag}.
While effective for task-specific benchmarks, their heavy external dependencies and fixed tool repertoires may limit generalization across diverse long-video scenarios.
More recent works explore lightweight internal retrieval mechanisms, where MLLMs output textual timestamps or structured queries to guide segment extraction \cite{xue2025omni, videomtr, longvt}.

Although more tightly integrated, these methods depend heavily on the temporal grounding capability of the underlying MLLMs.
As illustrated in Fig.~\ref{fig:tool_compares}, timestamp-based internal retrieval often entangles localization prediction with reasoning, making performance sensitive to the intrinsic temporal limitations of base models.

\section{Methodology}
In this section, we first present an overview of VideoTIR in Section \ref{subsec:overview}. Then we introduce the design of the multi-modal hierarchical toolkits in Section \ref{subsec:toolkits}. Section \ref{subsec:rls} focuses on the Zero-RL procedure with respect to the advantage estimator and rewards design. Section \ref{subsec:datasynth} present a general framework to synthesize high-quality trajectories from normal video QA-data or even pure video data.

\subsection{Overview}
\label{subsec:overview}
Different from MLLMs that process video-QA pairs within a single turn to provide the answer, human usually adapt a coarse-to-fine manner to divide the task in several sequential steps and then conquer by hierarchically adding fine-grained visual clues to decide whether or not to give the accurate answer. Therefore, as illustrated in Fig. \ref{fig:framework}, VideoTIR adopts a multi-turn manner to deal with the users' input videos and question. In considering of efficiency, videos are parsed with low resolution and fps initially and VideoTIR calls the textual router tool to decide whether the current visual information is enough to conduct an answer, or the router of using what tools to fetch the missing visual information corresponding to the question. In the latter cases, the fine-grained visual information generated by tools is combined with the former contexts and again parsed to the textual router tool. Such loops stop if the model conducts an answer or reaches the predefined number of max turns.

\subsection{Multi-Modal Hierarchical Retrieval Toolkits}
\label{subsec:toolkits}


\begin{figure}[t] 
  \centering
  \includegraphics[width=\textwidth]{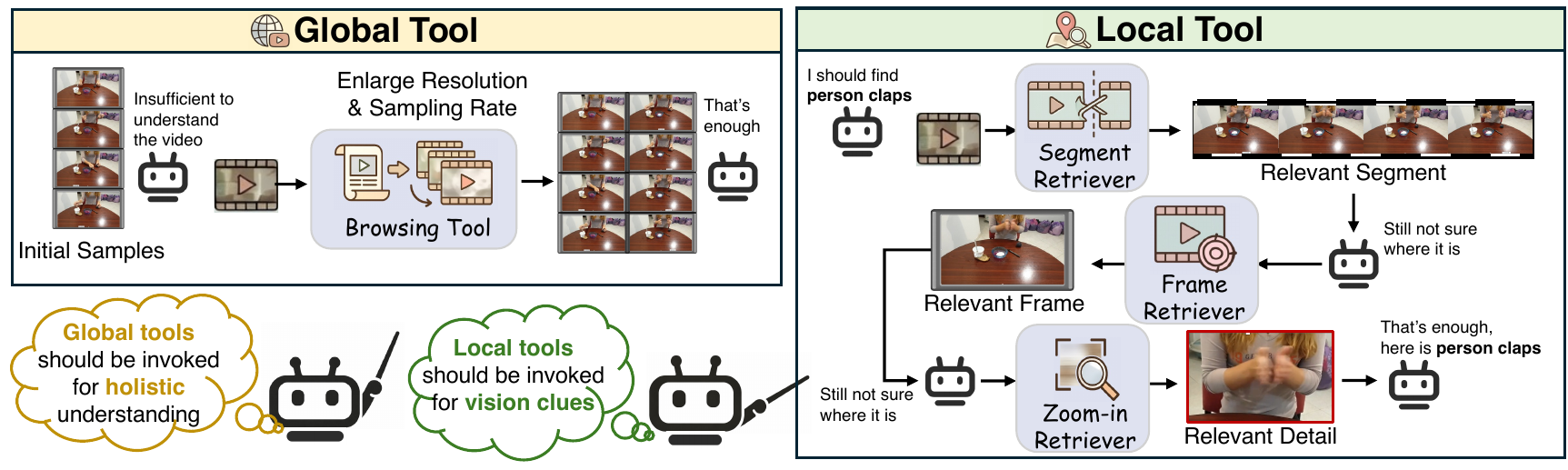}
  \caption{Hierarchical Visual Toolkits containing both Global and Local Tools. When there's a need for more information, the textual router calls global-level browsing tools for the general questions and detail-level tools for the questions targeting at finer perception of the videos.}
  \label{fig:toolkit}
\end{figure}

\noindent\textbf{Textual Router.} Our multi-modal hierarchical toolkit is designed to mimic human efficient retrieval behaviors. Humans usually first read the question and grasp the intent, and then roughly browse the videos with such intents. Therefore, we design a textual router tool as illustrated in Fig. \ref{fig:toolkit} to pay attention to the textual information and decide the potential corresponding objects or scenarios to answer the question. Combined with the current visual information, the router takes different behavior in different situations:
\begin{itemize}
    \item \textbf{Provide answers} when the corresponding visual information is enough.
    \item Otherwise, \textbf{Call browsing tools} when the questions' intents are at a global understanding of the videos and \textbf{temporal-spatial grounding tools} when the intents target at specific visual clues.
\end{itemize}

\noindent\textbf{Browsing Tool.} When the question aims at an entire understanding (\textit{e.g.}, what is the whole video about?) and the current visual information is not able to provide an answer, humans usually tend to gather more temporal and spatial information simultaneously by slower browsing and more frame-level attention. Therefore, our browsing tool simulates such a manner by enlarging both the resolution and the fps rate compared with before.

\noindent\textbf{Temporal-Spatial Grounding Tool.} When the questions need fine-grained visual clues (\textit{e.g.}, what happens between A and B?), humans usually need to locate the related parts of the videos and tend to ignore the other parts. Specifically, such locating behavior always performs a coarse-to-fine manner that first roughly locates temporally and then dives into the potential video segments for detailed spatial clues. Therefore, our temporal-spatial grounding tool is designed as a flexible hierarchical chains of retrievers with early stops:
\begin{itemize}
    \item \textbf{Segment Retriever.} Based on the text search results planned by the textual router tool, internal tools are called to locate relevant segments in the original video by selecting the video tokens with cosine similarity comparing to textual query tokens generated by MLLMs.
    \item \textbf{Frame Retriever.} Input a video segment, and this tool will retrieve the keyframes in that segment that are most relevant to the question and the retrieving sentence.
    \item \textbf{Zoom-in Retriever.} Input the image modality, and the tool will crop out the most relevant area to answer.
\end{itemize}
Details of the toolkit are in the Supplementary Materials.

\begin{figure}[t]
  \centering
  \includegraphics[width=0.96\linewidth]{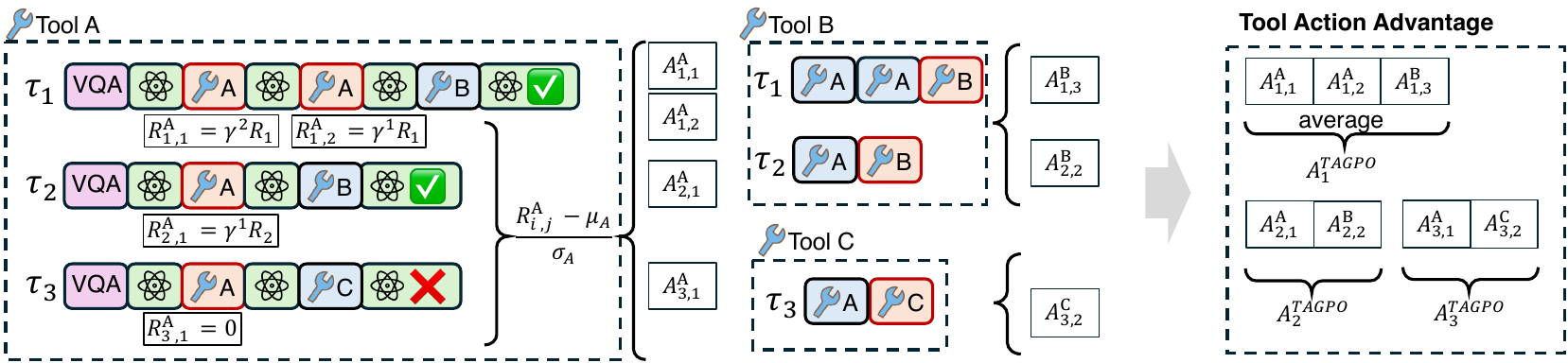}
  \caption{Visualization of Tool Action Advantage. We define rewards for each tool-calling action to penalize redundancy. The toolkit action advantage is the average of the tool advantages.}
  \label{fig:tagpo}
\end{figure}

\subsection{RL Algorithm Design}
\label{subsec:rls}

\begin{figure*}[!t] 
  \centering
  \includegraphics[width=1\linewidth]{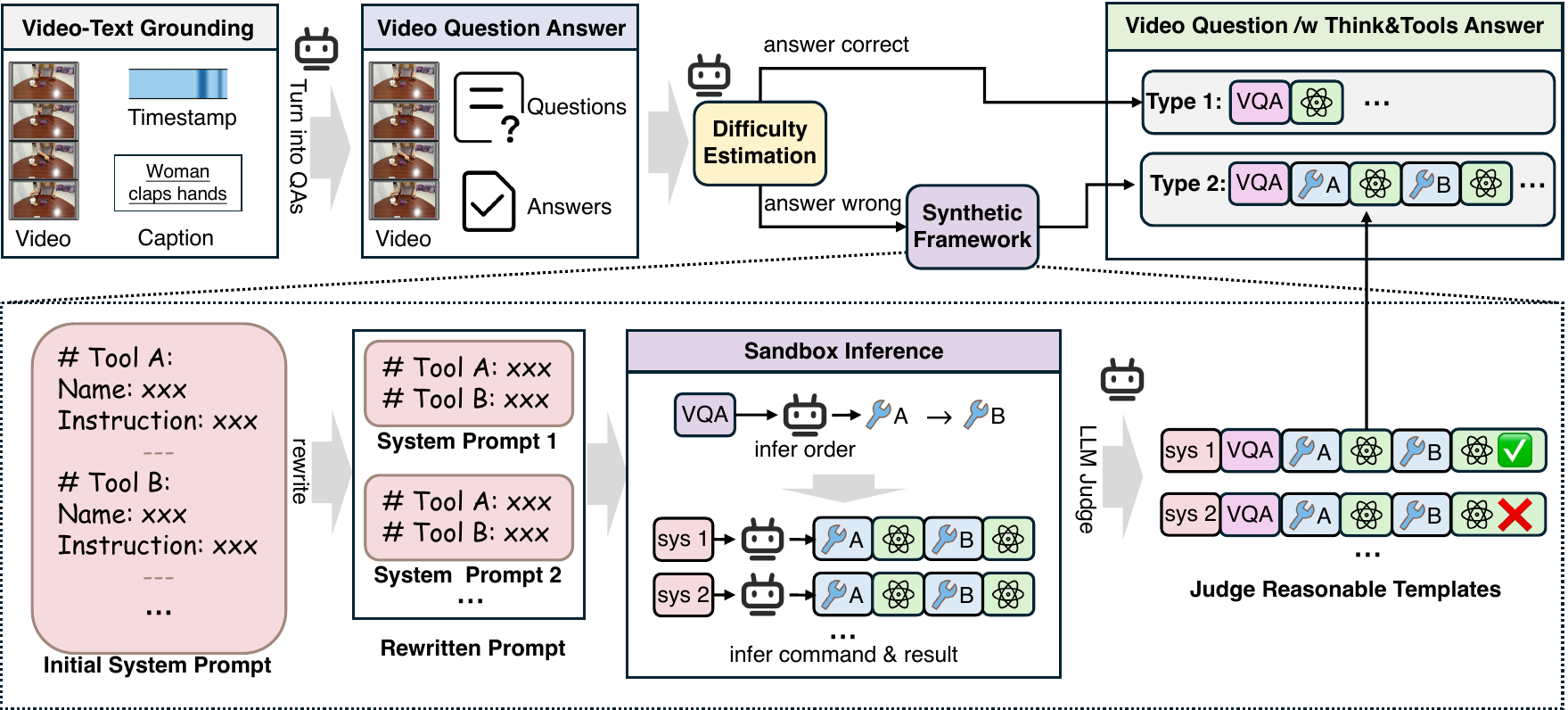}
  \caption{Framework of Data Synthesis. For video-text grounding datasets, we first convert them into QA datasets. Then, for the easy question, we synthesize trajectories that answer directly with no tool calls. For hard questions that model answers wrong, they are processed through a sandbox to generate tool calling trajectories. Finally, a large LLM is used to judge the rationality of the trajectories and we only keep the reasonable ones.}
  \label{fig:data_synth}
\end{figure*}

We denote the model‑generated sequence as $\tau_i$, $i\in \{1,2,...,N\}$, N is the rollout time. The episode‑level (global) reward comprises three components: an accuracy reward $R_{\mathrm{acc}}(\tau_i)=1$ if the final answer is correct; a format reward $R_{\mathrm{fmt}}(\tau_i)=1$ if the tool‑call string is parsable; and a tool‑use bonus  $\mathbb{I}_{\,R_{\mathrm{acc}}(\tau_i)>0}\cdot R_{\mathrm{tool}}(\tau_i)$ to encourage appropriate tool invocation \cite{deepeyes}. To be specific:

\begin{equation}
R_i := R_{\mathrm{acc}}(\tau_i) + R_{\mathrm{fmt}}(\tau_i)
+ \mathbb{I}_{R_{\mathrm{acc}}(\tau_i) > 0}\, R_{\mathrm{tool}}(\tau_i),
\quad
A_i^{\mathrm{GRPO}} := \frac{R_i - \mu(R)}{\sigma(R)}.
\end{equation}

where $R_i$ denotes the episode reward of $\tau_i$, $A^{GRPO}_i$ is the batch normalized episode advantage, and $\mu(R_{(\cdot)}),\sigma(R_{(\cdot)})$ are the mean and standard deviation of rewards computed over the N rollouts.

\noindent\textbf{Toolkit Action Grouped Policy Optimization.}
To assess the conciseness and precision of tool use, we move beyond episode‑level credit assignment and evaluate each tool invocation. For every call, we define a local reward guided by a simple principle: penalize redundancy in successful episodes while permitting exploration in failed episodes. We then compute a per‑call advantage, such that redundant invocations receive negative advantage whereas exploratory calls in failed rollouts receive null advantage. The trajectory level toolkit action advantage is obtained by averaging these per‑call advantages across all invocations in the trajectory. The resulting advantage allocation is illustrated in \cref{fig:tagpo}.

Assume that $\tau_i=(t_i^1,t_i^2,...,t_i^{L_i})$, where $t_i^j\in\{\mathrm{tool}_1,\mathrm{tool}_2,...\mathrm{tool}_K\}$ denotes a tool invocation, K is the number of available tools, $L_i$ s the number of tool calls in $\tau_i$. We define the per‑call reward for invoking $\mathrm{tool}_k$ at the $j_{th}$ step in trajectory $\tau_i$ as :
\begin{equation}\label{equation:tool_reward}
    R^{k}_{i,j} := \mathbb{I}_{\,R_{\mathrm{acc}}(\tau_i)>0}\cdot \gamma^{L_i-j} R_i
\end{equation}
where i,j can only take the indices at which $\mathrm{tool}_k$ is called. $\gamma\in(0,1]$ is a predefined decay coefficient.
This yields intuitive behavior in comparative cases:
\begin{itemize}
    \item If two successful trajectories (A, A, B) and (A, B) receive the same episode reward, the second A in (A, A, B) obtains the same per‑call reward as the single A in (A, B); the earlier, redundant A in (A, A, B) receives a smaller reward due to the additional decay, and therefore a negative advantage under intra‑tool normalization. This punishes the \textbf{overuse}.
    \item If (A, A, B) succeeds but (A, B) fails, all calls in (A, A, B) receive positive credit while all calls in (A, B) receive zero, reinforcing the necessity of the repeated A.
    \item If both (A, A, B) and (A, B) fail, no per‑call credit is assigned, yielding no relative advantage and thereby encouraging broader exploration of other tool combinations. This punishes the \textbf{misuse}.
\end{itemize}
The overall toolkit action advantage of $\tau_i$ is calculated as follows:
\begin{equation}
A^{k}_{i,j} := \frac{R^{k}_{i,j}-\mu(R^k_{(\cdot,\cdot)})}{\sigma(R^k_{(\cdot,\cdot)})}
\quad
A_i^{\text{TAGPO}} := \mu\!\left(A^{(\cdot)}_{i,(\cdot)}\right)
\end{equation}
where $A^{k}_{i,j}$ denotes the advantage of the invocation of $\mathrm{tool}_k$ at $j_{th}$ step in $\tau_i$, 
$\mu(R^k_{(\cdot,\cdot)}),\sigma(R^k_{(\cdot,\cdot)})$ is the mean and standard deviation of all the rewards of calling $\mathrm{tool}_k$, $A_i^{TAGPO}$ is the mean of the per‑call advantages for the trajectory.

We update the policy using a composite advantage:
\begin{equation}
\nabla_\theta \mathcal L =-\sum_{i} (A^{GRPO}_i+A^{TAGPO}_i)\nabla_\theta \log\pi_\theta(a_i|s_i)
\end{equation}
where $a_i$ denotes the action (output tokens) of the base model, $ s_i$ denotes the environment state.

\subsection{Pre-Rollout and Trajectory Synthesis\label{subsec:datasynth}}

Most video‑QA datasets provide only question–answer pairs and ground‑truth responses, lacking fine‑grained annotations about tool usage or intermediate reasoning. Consequently, during RL training the reward signal is largely restricted to answer accuracy and call‑string parseability, offering little supervision on the appropriateness of tool usage. To address this issue, as illustrated in Fig. \ref{fig:data_synth}, we introduce a multi‑stage trajectory synthesis framework.

We leverage an external MLLM (e.g., GLM‑4.5V) to synthesize both tool invocation sequences and pseudo environment feedback:
\begin{enumerate}
\item \textbf{Tool necessity filtering.} Using available video–text grounding, we generate multiple QA instances and prompt the MLLM to answer directly. Items solvable without tools are put aside.
\item \textbf{Tool order prediction.} For remaining items, we supply the VQA and the initial system prompt with toolkit template, and ask the MLLM to predict a plausible sequence of tool invocations.
\item \textbf{System prompt rewriting.} We rewrite the system prompt —especially the toolkit template —to diversify calling syntax and reasoning behaviors.
\item \textbf{Trajectory generation.} Conditioned on the rewritten prompt and the predicted tool order, the MLLM produces stepwise reasoning traces, explicit invocation commands, and expected environment feedback at each step within a sandboxed simulator.
\item \textbf{Trajectory adjudication.} The MLLM then ranks candidate trajectories by conciseness and precision, enabling selection of high‑quality exemplars.
\end{enumerate}



\section{Experiments}

\subsection{Setups}

\noindent\textbf{Benchmarks.}
We evaluated our method on three video understanding benchmarks, encompassing videos of varying lengths and capability systems.
\begin{itemize}
    \item \textbf{MVBench.} This benchmark is built from 11 short or medium duration video datasets containing STAR \cite{star}, PAXION \cite{paxion}, CLEVRER \cite{clevrer}, and \textit{etc} and evaluate the spatial-temporal reasoning ability by 20 diverse designed tasks. We adopt the full dataset of 4500 samples.
    \item \textbf{Video-MME.} Video-MME is a comprehensive benchmark for evaluating multimodal large models for video understanding. It covers 6 major visual domains and 30 subcategories, with video lengths ranging from 11 seconds to 1 hour. In addition to frame information, it also includes subtitles and audio to comprehensively examine the model's cross-modal and long-term temporal understanding capabilities. We adopt the whole 2700 samples for evaluation.
    \item \textbf{LongVideoBench.} This benchmark focuses on evaluating models' ability of event perception and relation understanding on medium or long videos. We adopt the validation set of this benchmark that contains 1377 samples.
\end{itemize}

\noindent\textbf{Baseline Models.}
We adopt \textbf{Qwen2.5-VL} \cite{qwen25} as our base models for post-training.
Specifically, we conduct experiments on two model scales, Qwen2.5-VL-3B and Qwen2.5-VL-7B, to evaluate the effectiveness and scalability of our method across different parameter regimes.
Qwen2.5-VL is a strong open-source vision-language model that supports both image and video inputs and demonstrates competitive performance on a wide range of multimodal understanding benchmarks.
Therefore, we focus on Qwen2.5-VL rather than earlier Qwen-VL-2 or more recent Qwen-VL-3 variants to ensure architectural consistency and controlled comparisons with other relative methods. 

\noindent\textbf{Details of Training and Inference.}
We use \textit{LLaMaFactory} \cite{llamafactory} as the training framework to Normal-SFT our 3B model with 8 NVIDIA 4090s and Random-Noising-SFT the 3B model with 2 AMD MI308Xs. In the agent-RL training stage, we use 8 NVIDIA H20s and \textit{Volcano Engine Reinforcement Learning} (VeRL) as the training framework to train the 7B Zero-RL model and use 4 AMD MI308Xs to train the 3B Zero-RL and Cold-Started models. 

We set the number of max turns as 4 and rollout per sample as 8. Also, we following the default max frames as 32 per video and max pixels of the initial turn per frame as $14 * 14 * 28 * 28$ (196 tokens), and total pixels as $10 * 14 * 14 * 28 * 28$ (1960 tokens). We adopt different coefficient of the rewards between zero-RL settings and SFT cold-start settings. This is motivated by observation that the ability of following formality instruction after SFT is at a high initial range and therefore the model should learn more from other rewards.
We also adopt online rollout filtering \cite{simpletir} to ensure training stability.

While evaluating, the frame resolution settings "Low" in Tab.~\ref{tab:main_table} are the same as training (1960 video tokens) and "high" refers to 1 fps sampling (at max 32 frames and 16384 video tokens). All the evaluations are 8 NVIDIA 4090s or 4 AMD MI308Xs.

\begin{figure}[t]
  \centering
  \includegraphics[width=\columnwidth]{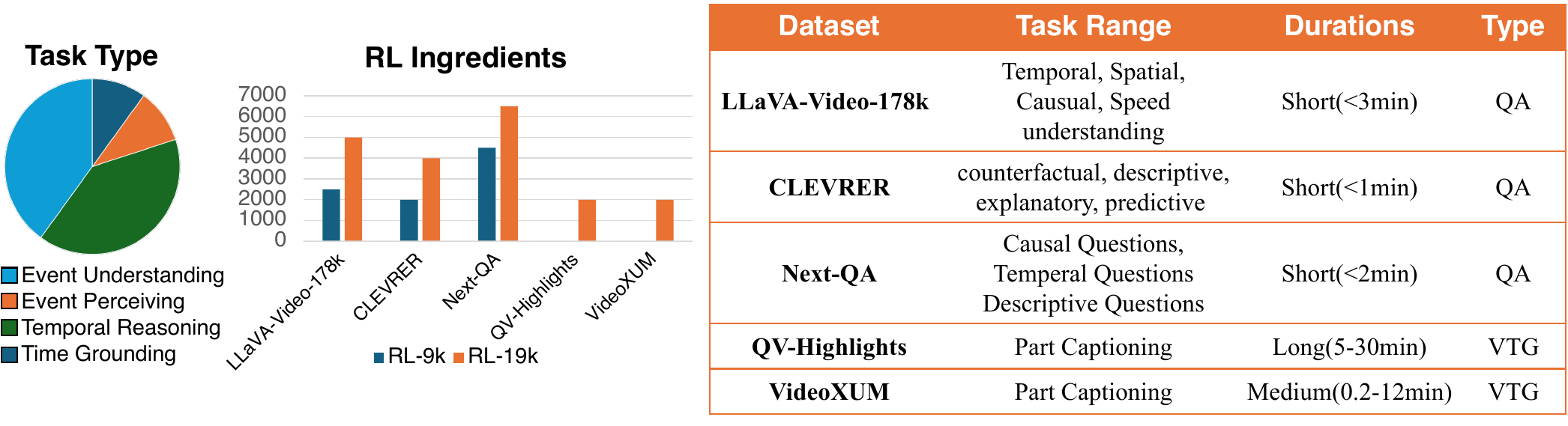}
  \caption{Distribution and Task Range of Curated Datasets (VideoSIAM is not included in). We selected 4 general tasks that potentially need tools for finer perception. We also synthesize high-quality trajectories as the SFT dataset for 3B model cold starting.}
  \label{fig:data_ingredient}
\end{figure}

\noindent\textbf{RL Dataset Preparation and Curation.}
As shown in Fig. \ref{fig:data_ingredient}, we curated our dataset from the LLaVa-Video-155k, CLEVRER, PerceptionTest, and VideoSIAM. The first 3 datasets are curated by difficulty estimation and ground-truth verification, that do 10 times of examination and select the medium difficulty samples that answer correctly in range (3,7). For 3B model, we select 4.5k RL training samples and for 7B model 9k samples are selected.

\noindent\textbf{SFT Trajectory Synthesis and Curation.}
We synthesize high-quality trajectories from video-QA datasets NextQA \cite{xiao2021next} and NextGQA \cite{nextgqa}, Perceptiontest and VTG datasets VideoXUM \cite{videoxum}, QV-Highlights \cite{qvhilights}. We adopt 4.5k of the tool-invoking trajectories combine with 3 times more textual CoT data from VideoR1 \cite{videor1} as the cold start SFT dataset (17,384 in total). We do not apply extra difficulty estimation since this procedure has already done in the synthetic framework. The detailed ingredients are shown in Fig. \ref{fig:data_ingredient} more details of the synthetic framework are in the supplementary materials.

\subsection{Main Results}
\begin{table}[t]
\centering
\caption{Main Results of the comparison and our method. Our method surpasses the base Qwen2.5-VL-7B model and many comparison methods that require more frames and higher resolution (Res.) as inputs. \\
$\dagger$ Following \cite{zhong2025rethinking}, we observe that enabling thinking mode in small MLLMs (<7B) degrades LVU performance. For fairness, we also report thinking-mode results.  \\
$\ddagger$ These methods are SFTed on approximately $100$k TIR and non-TIR trajectories.} 
\small
\label{tab:main_table}

\resizebox{\textwidth}{!}{%
\begin{tabular}{l c c c c c c c c c c}
\toprule

\multirow{2}{*}{\textbf{Model}} &
\multirow{2}{*}{\textbf{Size}} &
\multirow{2}{*}{\textbf{Input}} &
\multirow{2}{*}{\textbf{Res.}} &
\multirow{2}{*}{\textbf{Thinking}} &
\multicolumn{4}{c}{\textbf{V$\text{-}$MME$_{\text{w/o sub}}$}} &
\multirow{2}{*}{\textbf{MVBench}} &
\multirow{2}{*}{\textbf{LVB$_{\text{val}}$}} \\
\cmidrule(lr){6-9}
 & & & & & \textbf{Short} & \textbf{Medium} & \textbf{Long} & \textbf{Mean} & & \\
\midrule
InternVL3\cite{internvl} 
& 2B & 16-frame & High & $\times \dagger$ 
& 67.3 & 54.2 & 46.2 & 55.9 & 63.9 & - \\
\rowcolor{LightBlueRow}
Qwen2\text{-}VL\cite{wang2024qwen2} 
& 2B & 16-frame & High & $\times$ 
& 58.1 & 46.0 & 41.3 & 48.5 & 57.3 & 46.3 \\
Qwen2.5\text{-}VL\cite{qwen25} 
& 3B & 16-frame & High & $\times$ 
& 63.4 & 50.8 & 47.7 & 54.0 & 60.9 & \underline{51.8} \\



\midrule
\rowcolor{LightBlueRow}
Video\text{-}LLaVa\cite{zhang2023video} 
& 7B & 8-frame & High & $\checkmark \dagger$ 
& - & - & - & 39.9 & 34.1 & 39.1 \\
VideoChat2\cite{wang2025videochat} 
& 7B & 16-frame & High & $\checkmark$ 
& - & - & - & 39.2 & 51.1 & 39.3 \\
\rowcolor{LightBlueRow}
Video\text{-}R1\cite{videor1} 
& 7B & 16-frame & High & $\checkmark$ 
& - & - & - & 57.4 & \underline{62.7} & - \\
Video\text{-}XL\cite{video-xl} 
& 7B & 128-frame & High & $\checkmark$ 
& - & - & - & 55.5 & 55.3 & 50.7 \\
\rowcolor{LightBlueRow}
Video\text{-}MTR\cite{videomtr} 
& 7B & 32-frame & High & $\checkmark$ 
& - & - & - & \underline{59.0} & - & - \\
LongVT\text{-}RL $\ddagger$ \cite{longvt}
& 7B & 64-frame & High & $\checkmark$ 
& - & - & - & \textbf{66.1} & - & - \\
\rowcolor{LightBlueRow}
Qwen2.5\text{-}VL 
& 7B & 16-frame & High & $\times$ 
& 63.1 & 50.9 & 47.3 & 53.8 & 60.6 & \underline{51.8} \\

Qwen2.5\text{-}VL 
& 7B & 16-frame & High & $\checkmark$ 
& 61.9 & 46.7 & 39.8 & 49.5 & 55.2 & 47.5 \\

\rowcolor{LightBlueRow}
Qwen2.5\text{-}VL
& 7B & 8-frame & Low & $\checkmark$ 
& - & - & - & 35.0 & 40.3 & 28.1 \\

\rowcolor{OursHighlight}

w/ Tool (Zero-Shot) 
& 7B & 8-frame & Low & $\checkmark$ 
& 52.2 & 46.4 & 30.1 & 42.9 & 47.8 & 38.6 \\

\rowcolor{OursHighlight}

w/ Tool + GRPO 
& 7B & 8-frame & Low & $\checkmark$ 
& 64.3 & 49.7 & 48.3 & 54.1 & 56.9 & 50.9 \\

\rowcolor{OursHighlight}

w/ Tool + TAGPO 
& 7B & 8-frame & Low & $\checkmark$ 
& 65.1 & 50.4 & 47.1 & 54.2 & 56.0 & 51.3 \\

\rowcolor{OursHighlight}

w/ Tool + TAGPO 
& 7B & 16-frame & High & $\checkmark$ 
& 67.2  & 55.8  & 50.7  & 57.9   & \textbf{64.3} & \textbf{53.1} \\

\bottomrule
\end{tabular}
}
\end{table}

\noindent\textbf{Short- and Medium-form Video Understanding.}
On short or medium duration videos, our methods surpass the base model Qwen2.5-VL-7B. Compared with VideoMTR, the increasing performance on MVBench and LongVideoBench illustrates that external efficient tools can achieve better visual information acquisition.

\noindent\textbf{Long-form Video Understanding.}
Surprisingly, our model shows more increase on these long form videos.  Also, the performance of VideoMTR is constrained since such constrained sampling rates lost too much temporal information, leading to difficulties retrieving clues between the sparse sampling frames by only MLLMs themselves. By contrast, our model uses efficient lightweight internal tools and is able to capture critical clues even under a sparse sampling setting.



\begin{figure}[t]
  \centering
  \begin{subfigure}[t]{0.40\columnwidth}
    \centering
    \includegraphics[width=\linewidth]{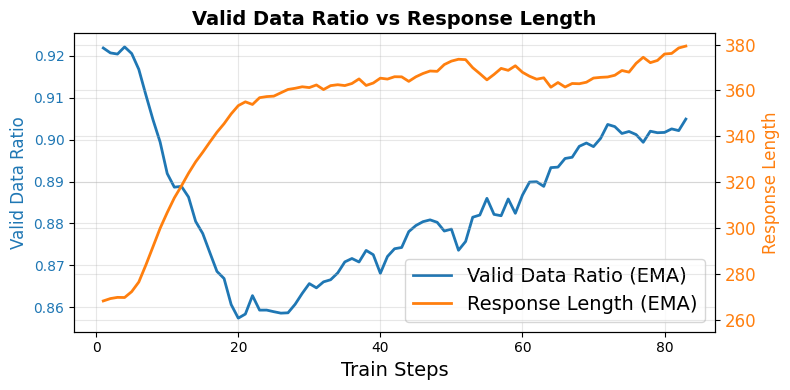}
    \caption{Valid data ratio vs. response length.}
    \label{fig:valid_data_ratio}
  \end{subfigure}
  \hfill
  \begin{subfigure}[t]{0.49\columnwidth}
    \centering
    \includegraphics[width=\linewidth]{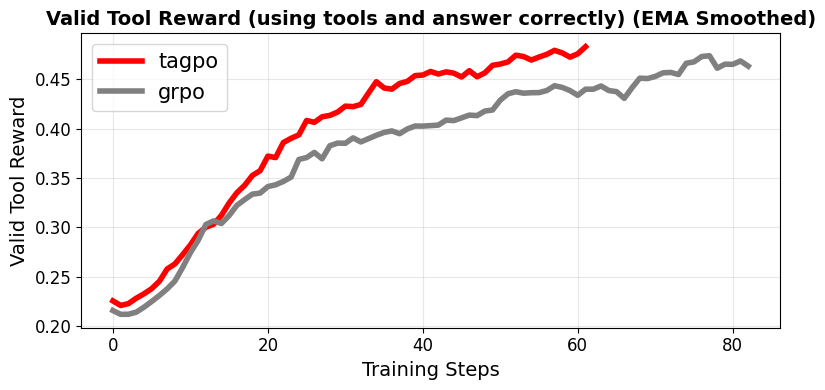}
    \caption{TAGPO vs. episode-level GRPO.}
    \label{fig:tagpo_vs_grpo}
  \end{subfigure}

  \caption{
  Training dynamics analysis.
  (a) At early stages, response length increases while format quality drops,
  indicating the model prioritizes rational tool exploration.
  Once response length stabilizes, format quality gradually improves,
  suggesting a balance between rationality and formality.
  (b) TAGPO accelerates valid tool learning.
  The valid tool reward rises significantly faster than episode-level GRPO,
  reducing the steps required to exceed 0.45 by approximately 50\%.
  }
  \label{fig:training_dynamics}
\end{figure}

\noindent\textbf{Cases Analysis.}
We show several whole trajectories $\tau_{ori}$, including both the textual and visual modality from the evaluation benchmarks. We find that the formality ability gained from the SFT stage is robust enough that even lower format reward in the sequential RL stage can be kept (also see the format quality in Fig. \ref{fig:valid_data_ratio}). Moreover, as the training step increases, the tool calling is more proper. For example, at the early stage, the model is likely to generate retrieving sentences that are too short for the tools. After RL, the generated retrieving sentences are shorter and proper for effectively calling the toolkit. The increasing "valid tool call" curve (Fig. \ref{fig:tagpo}) and the curve of response length (Fig. \ref{fig:valid_data_ratio}) of the training stage also indicate this fact.

\subsection{Efficiency Analysis}

\begin{table}[t]
\centering
\begin{minipage}[t]{0.36\columnwidth}
\centering
\caption{
Formatting quality under different prompts and training strategies.
The 3B base model produces near-zero valid formatting, while format-oriented SFT restores compliance and enables RL.
}
\label{tab:why_3b_should_sft}

\small
\resizebox{\linewidth}{!}{%
\begin{tabular}{l c c}
\toprule
\textbf{Model} & \textbf{Sys. Pr.} & \textbf{Fmt. Qual.} \\
\midrule
Qwen2.5-VL-7B & Short & $\leq$1.0 \\
\rowcolor{LightBlueRow}
Qwen2.5-VL-7B & Long & 81.7 \\
Qwen2.5-VL-3B & Short & $\leq$1.0 \\
\rowcolor{LightBlueRow}
w/ SFT & Short & 91.7 \\
Qwen2.5-VL-3B & Long & $\leq$1.0 \\
\rowcolor{LightBlueRow}
w/ SFT & Long & 90.1 \\
\bottomrule
\end{tabular}%
}
\end{minipage}\hfill
\begin{minipage}[t]{0.61\columnwidth}
\centering
\caption{
Efficiency comparison of tool-integrated reasoning designs.
VideoTIR's internal neural-network toolkit avoids external API or specialist-model calls, yielding the fewest average tool invocations (1.42) and the lowest latency (4.6s).
Latency is measured end-to-end excluding network overhead; $^\dagger$ values are manually set by the original papers.
}
\label{tab:efficiency}

\footnotesize
\setlength{\tabcolsep}{3pt}
\resizebox{\linewidth}{!}{%
\begin{tabular}{lcccc}
\toprule
\textbf{Metrics} & \textbf{VideoMA}\cite{kugo2025videomultiagents} & \textbf{COLT}\cite{liu2026colt} & \textbf{VideoMTR}\cite{videomtr} & \textbf{VideoTIR} \\
\midrule
Tools Type & Ext.\,(API) & Ext.\,(tools) & Internal & Internal \\
Avg.\ Tool Invocations & $3^\dagger$ & $3^\dagger$ & $2^\dagger$ & \textbf{1.42} \\
Latency ($\downarrow$) & 27.4s & 68.3s & 5.9s & \textbf{4.6s} \\
\bottomrule
\end{tabular}%
}

{\footnotesize VideoMA issues multiple GPT-4o/Gemini API calls (API-bound); COLT uses 8 external specialist models.}
\end{minipage}
\end{table}

Beyond accuracy, toolkit design directly affects inference efficiency. Existing methods depend on costly external APIs or specialist models~\cite{kugo2025videomultiagents,liu2026colt}, and even lightweight internal retrieval in VideoMTR~\cite{videomtr} requires repeated calls. As shown in Tab.~\ref{tab:efficiency}, these methods need 2--3 invocations and 5.9--68.3s latency. VideoTIR instead uses a lightweight internal toolkit built on the backbone encoders, avoiding inter-process overhead and enabling GPU parallelization. 

\subsection{Ablation Studies}

\noindent\textbf{Synthetic Trajectories in SFT.}
Besides the quantitative results in Tab. \ref{tab:main_table} that indicates the performance gains after the SFT procedure, we do detailed experiments as shown in Tab. \ref{tab:why_3b_should_sft}. 

We attempt to implement Zero-RL on 3B but fails because the complex format instruction is not suitable for the 3B model, as the valid tool call strings are extremely rare and thus the advantage estimator fails since the in-group std of the advantages is near 0.
Moreover, we analyze the failure string patterns of the 3B model and design a series of corrective strategies to address common erroneous JSON formats (see Supp.). 

\begin{table}[t]
\centering

\caption{
Normal-SFT (N.SFT) vs. Random-SFT (R.SFT) on the 3B model after 300 steps (batch size = 32).
Both achieve near-perfect formatting, while Random-SFT yields higher validation accuracy.
}
\label{tab:sft_compare}

\small
\resizebox{0.85\columnwidth}{!}{%
\begin{tabular}{lcccc}
\toprule
 & \textbf{MvBench*} & \textbf{VideoMME*} & \textbf{LvBench*} & \textbf{Format Quality} \\
\midrule
N.SFT & 36.3 & 35.0 & 34.0 & $\sim 100$ \\
\rowcolor{LightBlueRow}
R.SFT $\dagger$ & 40.7 & 38.5 & 39.0 & $\sim 100$ \\
\bottomrule
\end{tabular}%
}

{\footnotesize $*$ Randomly sampled 10\%.}

\end{table}

We further compare Normal SFT and Random Noised SFT on MVBench. Surprisingly (Tab.~\ref{tab:sft_compare}), on our dataset of approximately 4.5k samples, Random-Noised SFT, where the vision modality is replaced with placeholders and masked at the attention level (see supplementary materials), outperforms Normal-SFT while maintaining high Format Quality (Detailed discussions can be found in the Supps.).

\begin{table}[t]
\centering

\begin{minipage}[t]{0.52\columnwidth}
\centering
\setlength{\tabcolsep}{5pt}
\renewcommand{\arraystretch}{0.85}
\caption{Reasonability analysis of the textual router on Video-MME.
For Information Synopsis (I.S.) tasks, which require global video understanding,
the router predominantly selects browsing-based tool chains.
In contrast, for Object Recognition (O.R.) tasks that focus on specific local regions,
the router favors grounding-oriented tool chains.}
\label{tab:ispr_analysis}
\begin{tabularx}{\linewidth}{l X X}
\toprule
\textbf{Tool Chain Type} & \textbf{I.S.} & \textbf{O.R.} \\
\midrule
\rowcolor{LightBlueRow}
Browser                 & 309 & 3 \\
Retriever \& followings & 1   & 335 \\
\rowcolor{LightBlueRow}
Direct Answer           & 13  & 16 \\
\bottomrule
\end{tabularx}
\end{minipage}
\hfill
\begin{minipage}[t]{0.46\columnwidth}
\centering
\setlength{\tabcolsep}{4pt}
\renewcommand{\arraystretch}{1.1}
\small
\caption{Ablation study of TAGPO on the 7B model.
Validation accuracy is reported at 30 training steps
with a batch size of 32 for both GRPO and TAGPO.
TAGPO achieves consistently higher performance,
demonstrating improved optimization efficiency in early-stage training.}
\label{tab:tagpo_ablation}
\begin{tabular}{lc}
\toprule
\textbf{Model} & \textbf{Val Acc.} \\
\midrule
7B w/ GRPO  & 21.1 \\
\rowcolor{LightBlueRow}
7B w/ TAGPO & 24.6 \\
\bottomrule
\end{tabular}
\end{minipage}

\end{table}


\noindent\textbf{Property of Global Tool and Retrieval Tools.}
We select 2 specific tasks "Information Synopsis" and "Percept \& Recognize" in the VideoMME benchmark and analysis the tool callings on these samples. Information Synopsis needs a global understanding of the video and is suitable for "Browsing Tool" while the Percept \& Recognize task needs to accurately recall the video segments and is suitable for the temporal-spatial grounding tools. The results in Tab. \ref{tab:ispr_analysis} shows that our textual router tool is able to make the proper decisions.



\noindent\textbf{TAGPO.}
We evaluate the quality of tool calling from the perspectives of both misuse and overuse.
To measure misuse, we adopt the valid tool reward, defined as the ratio of tool invocations that lead to correct answers. In addition, we compute the average number of tool calls per trajectory at each training step.
As shown in Fig.~\ref{fig:tagpo_vs_grpo}, TAGPO consistently surpasses the baseline GRPO in terms of valid tool reward, while maintaining a comparable level of tool invocation (i.e., tool reward) during the early training stage.
From the overall growth trend of the final accuracy reward, TAGPO accelerates early policy optimization by reducing unnecessary exploration time.

For the accuracy metric, as shown in Tab. \ref{tab:tagpo_ablation} before, we also evaluate the 30-step accuracy of GRPO and TAGPO and find that the latter method reaches a higher accuracy on validation settings. 


\section{Conclusion}
We present VideoTIR, a multi-turn, multi-tool reasoning framework designed to accurately and efficiently locate the information required to solve long video understanding.
For 7B models, we employ end-to-end RL to learn effective tool usage strategies.
We further propose a Tool Action Group Policy Optimization (TAGPO) method, which computes advantages across tool calls within the same category. This approach provides finer-grained rewards for rational tool usage.
To address this problem, we develop a sandbox-based trajectory synthesis framework to construct high-quality datasets for both SFT and RL training.

\section*{Acknowledgements}
This work was supported in part by the National Natural Science Foundation of China, under Grant Nos. 62192783, 62276128, and 62406140; the Young Elite Scientists Sponsorship Program by China Association for Science and Technology, under Grant No. 2023QNRC001; the Key Research and Development Program of Jiangsu Province, under Grant No. BE2023019; and the Jiangsu Natural Science Foundation, under Grant Nos. BK20221441 and BK20241200. The authors would like to thank the support of Huawei Ascend Cloud Ecological Development Project.

\bibliographystyle{splncs04}
\bibliography{main}

@String(CVPR= {IEEE Conf. Comput. Vis. Pattern Recog.})

@String(ICCV= {Int. Conf. Comput. Vis.})

@String(ECCV= {Eur. Conf. Comput. Vis.})

@String(ICLR = {Int. Conf. Learn. Represent.})

@String(CVPR  = {CVPR})

@String(ICCV  = {ICCV})

@String(ECCV  = {ECCV})

@String(ICLR  = {ICLR})

@article{videomind,
  title={VideoMind: A Chain-of-LoRA Agent for Long Video Reasoning},
  author={Liu, Ye and Lin, Kevin Qinghong and Chen, Chang Wen and Shou, Mike Zheng},
  journal={arXiv},
  year={2025}
}

@article{star,
  title={Star: A benchmark for situated reasoning in real-world videos},
  author={Wu, Bo and Yu, Shoubin and Chen, Zhenfang and Tenenbaum, Joshua B and Gan, Chuang},
  journal={arXiv},
  year={2024}
}

@article{paxion,
  title={Paxion: Patching action knowledge in video-language foundation models},
  author={Wang, Zhenhailong and Blume, Ansel and Li, Sha and Liu, Genglin and Cho, Jaemin and Tang, Zineng and Bansal, Mohit and Ji, Heng},
  journal={NeurIPS},
  year={2023}
}

@article{llamafactory,
  title={Llamafactory: Unified efficient fine-tuning of 100+ language models},
  author={Zheng, Yaowei and Zhang, Richong and Zhang, Junhao and Ye, Yanhan and Luo, Zheyan and Feng, Zhangchi and Ma, Yongqiang},
  journal={arXiv},
  year={2024}
}

@article{videoxum,
  title={Videoxum: Cross-modal visual and textural summarization of videos},
  author={Lin, Jingyang and Hua, Hang and Chen, Ming and Li, Yikang and Hsiao, Jenhao and Ho, Chiuman and Luo, Jiebo},
  journal={IEEE Trans. Multimed.},
  year={2023},
}

@article{simpletir,
  title={Simpletir: End-to-end reinforcement learning for multi-turn tool-integrated reasoning},
  author={Xue, Zhenghai and Zheng, Longtao and Liu, Qian and Li, Yingru and Zheng, Xiaosen and Ma, Zejun and An, Bo},
  journal={arXiv},
  year={2025}
}

@article{qvhilights,
  title={Detecting moments and highlights in videos via natural language queries},
  author={Lei, Jie and Berg, Tamara L and Bansal, Mohit},
  journal={NeurIPS},
  year={2021}
}

@article{clevrer,
  title={Clevrer: Collision events for video representation and reasoning},
  author={Yi, Kexin and Gan, Chuang and Li, Yunzhu and Kohli, Pushmeet and Wu, Jiajun and Torralba, Antonio and Tenenbaum, Joshua B},
  journal={arXiv},
  year={2019}
}

@article{videomtr,
  title={Video-mtr: Reinforced multi-turn reasoning for long video understanding},
  author={Xie, Yuan and Chen, Tianshui and Ge, Zheng and Ni, Lionel},
  journal={arXiv},
  year={2025}
}

@inproceedings{videoagent,
  title={Videoagent: Long-form video understanding with large language model as agent},
  author={Wang, Xiaohan and Zhang, Yuhui and Zohar, Orr and Yeung-Levy, Serena},
  booktitle={ECCV},
  year={2024},
}

@inproceedings{wang2025lvbench,
  title={Lvbench: An extreme long video understanding benchmark},
  author={Wang, Weihan and He, Zehai and Hong, Wenyi and Cheng, Yean and Zhang, Xiaohan and Qi, Ji and Ding, Ming and Gu, Xiaotao and Huang, Shiyu and Xu, Bin and others},
  booktitle={ICCV},
  year={2025}
}

@inproceedings{wu2019long,
  title={Long-term feature banks for detailed video understanding},
  author={Wu, Chao-Yuan and Feichtenhofer, Christoph and Fan, Haoqi and He, Kaiming and Krahenbuhl, Philipp and Girshick, Ross},
  booktitle={CVPR},
  year={2019}
}

@inproceedings{weng2024longvlm,
  title={Longvlm: Efficient long video understanding via large language models},
  author={Weng, Yuetian and Han, Mingfei and He, Haoyu and Chang, Xiaojun and Zhuang, Bohan},
  booktitle={ECCV},
  year={2024},
}

@article{qwen25,
  title={Qwen 2.5: A comprehensive review of the leading resource-efficient llm with potential to surpass all competitors},
  author={Ahmed, Imtiaz and Islam, Sadman and Datta, Partha Protim and Kabir, Imran and Chowdhury, Naseef Ur Rahman and Haque, Ahshanul},
  journal={Authorea Preprints},
  year={2025},
  publisher={Authorea}
}

@article{glm,
  title={Chatglm: A family of large language models from glm-130b to glm-4 all tools},
  author={GLM, Team and Zeng, Aohan and Xu, Bin and Wang, Bowen and Zhang, Chenhui and Yin, Da and Zhang, Dan and Rojas, Diego and Feng, Guanyu and Zhao, Hanlin and others},
  journal={arXiv},
  year={2024}
}

@article{videor1,
  title={Video-r1: Reinforcing video reasoning in mllms},
  author={Feng, Kaituo and Gong, Kaixiong and Li, Bohao and Guo, Zonghao and Wang, Yibing and Peng, Tianshuo and Wu, Junfei and Zhang, Xiaoying and Wang, Benyou and Yue, Xiangyu},
  journal={arXiv},
  year={2025}
}

@inproceedings{video-xl,
  title={Video-xl: Extra-long vision language model for hour-scale video understanding},
  author={Shu, Yan and Liu, Zheng and Zhang, Peitian and Qin, Minghao and Zhou, Junjie and Liang, Zhengyang and Huang, Tiejun and Zhao, Bo},
  booktitle={CVPR},
  year={2025}
}

@inproceedings{internvl,
  title={Internvl: Scaling up vision foundation models and aligning for generic visual-linguistic tasks},
  author={Chen, Zhe and Wu, Jiannan and Wang, Wenhai and Su, Weijie and Chen, Guo and Xing, Sen and Zhong, Muyan and Zhang, Qinglong and Zhu, Xizhou and Lu, Lewei and others},
  booktitle={CVPR},
  year={2024}
}

@article{thyme,
  title={Thyme: Think beyond images},
  author={Zhang, Yi-Fan and Lu, Xingyu and Yin, Shukang and Fu, Chaoyou and Chen, Wei and Hu, Xiao and Wen, Bin and Jiang, Kaiyu and Liu, Changyi and Zhang, Tianke and others},
  journal={arXiv},
  year={2025}
}

@inproceedings{wu2021towards,
  title={Towards long-form video understanding},
  author={Wu, Chao-Yuan and Krahenbuhl, Philipp},
  booktitle={CVPR},
  year={2021}
}

@article{tong2022videomae,
  title={Videomae: Masked autoencoders are data-efficient learners for self-supervised video pre-training},
  author={Tong, Zhan and Song, Yibing and Wang, Jue and Wang, Limin},
  journal={NeurIPS},
  year={2022}
}

@article{alayrac2022flamingo,
  title={Flamingo: a visual language model for few-shot learning},
  author={Alayrac, Jean-Baptiste and Donahue, Jeff and Luc, Pauline and Miech, Antoine and Barr, Iain and Hasson, Yana and Lenc, Karel and Mensch, Arthur and Millican, Katherine and Reynolds, Malcolm and others},
  journal={NeurIPS},
  year={2022}
}

@article{zhang2023video,
  title={Video-llama: An instruction-tuned audio-visual language model for video understanding},
  author={Zhang, Hang and Li, Xin and Bing, Lidong},
  journal={arXiv},
  year={2023}
}

@inproceedings{maaz2024video,
  title={Video-chatgpt: Towards detailed video understanding via large vision and language models},
  author={Maaz, Muhammad and Rasheed, Hanoona and Khan, Salman and Khan, Fahad},
  booktitle={ACL},
  year={2024}
}

@inproceedings{lin2024video,
  title={Video-llava: Learning united visual representation by alignment before projection},
  author={Lin, Bin and Ye, Yang and Zhu, Bin and Cui, Jiaxi and Ning, Munan and Jin, Peng and Yuan, Li},
  booktitle={EMNLP},
  year={2024}
}

@article{wang2024qwen2,
  title={Qwen2-vl: Enhancing vision-language model's perception of the world at any resolution},
  author={Wang, Peng and Bai, Shuai and Tan, Sinan and Wang, Shijie and Fan, Zhihao and Bai, Jinze and Chen, Keqin and Liu, Xuejing and Wang, Jialin and Ge, Wenbin and others},
  journal={arXiv},
  year={2024}
}

@inproceedings{li2024mvbench,
  title={Mvbench: A comprehensive multi-modal video understanding benchmark},
  author={Li, Kunchang and Wang, Yali and He, Yinan and Li, Yizhuo and Wang, Yi and Liu, Yi and Wang, Zun and Xu, Jilan and Chen, Guo and Luo, Ping and others},
  booktitle={CVPR},
  year={2024}
}

@inproceedings{fu2025video,
  title={Video-mme: The first-ever comprehensive evaluation benchmark of multi-modal llms in video analysis},
  author={Fu, Chaoyou and Dai, Yuhan and Luo, Yongdong and Li, Lei and Ren, Shuhuai and Zhang, Renrui and Wang, Zihan and Zhou, Chenyu and Shen, Yunhang and Zhang, Mengdan and others},
  booktitle={CVPR},
  year={2025}
}

@article{mangalam2023egoschema,
  title={Egoschema: A diagnostic benchmark for very long-form video language understanding},
  author={Mangalam, Karttikeya and Akshulakov, Raiymbek and Malik, Jitendra},
  journal={NeurIPS},
  year={2023}
}

@inproceedings{xiao2021next,
  title={Next-qa: Next phase of question-answering to explaining temporal actions},
  author={Xiao, Junbin and Shang, Xindi and Yao, Angela and Chua, Tat-Seng},
  booktitle={CVPR},
  year={2021}
}

@inproceedings{grauman2022ego4d,
  title={Ego4d: Around the world in 3,000 hours of egocentric video},
  author={Grauman, Kristen and Westbury, Andrew and Byrne, Eugene and Chavis, Zachary and Furnari, Antonino and Girdhar, Rohit and Hamburger, Jackson and Jiang, Hao and Liu, Miao and Liu, Xingyu and others},
  booktitle={CVPR},
  year={2022}
}

@inproceedings{yao2022react,
  title={React: Synergizing reasoning and acting in language models},
  author={Yao, Shunyu and Zhao, Jeffrey and Yu, Dian and Du, Nan and Shafran, Izhak and Narasimhan, Karthik R and Cao, Yuan},
  booktitle={ICLR},
  year={2022}
}

@article{shinn2023reflexion,
  title={Reflexion: Language agents with verbal reinforcement learning},
  author={Shinn, Noah and Cassano, Federico and Gopinath, Ashwin and Narasimhan, Karthik and Yao, Shunyu},
  journal={NeurIPS},
  year={2023}
}

@inproceedings{fan2024videoagent,
  title={Videoagent: A memory-augmented multimodal agent for video understanding},
  author={Fan, Yue and Ma, Xiaojian and Wu, Rujie and Du, Yuntao and Li, Jiaqi and Gao, Zhi and Li, Qing},
  booktitle={ECCV},
  year={2024},
}

@inproceedings{ma2025drvideo,
  title={Drvideo: Document retrieval based long video understanding},
  author={Ma, Ziyu and Gou, Chenhui and Shi, Hengcan and Sun, Bin and Li, Shutao and Rezatofighi, Hamid and Cai, Jianfei},
  booktitle={CVPR},
  year={2025}
}

@article{jeoung2024adaptive,
  title={Adaptive Video Understanding Agent: Enhancing efficiency with dynamic frame sampling and feedback-driven reasoning},
  author={Jeoung, Sullam and Huybrechts, Goeric and Ganesh, Bhavana and Galstyan, Aram and Bodapati, Sravan},
  journal={arXiv},
  year={2024}
}

@article{chen2025lvagent,
  title={Lvagent: Long video understanding by multi-round dynamical collaboration of mllm agents},
  author={Chen, Boyu and Yue, Zhengrong and Chen, Siran and Wang, Zikang and Liu, Yang and Li, Peng and Wang, Yali},
  journal={arXiv},
  year={2025}
}

@article{kugo2025videomultiagents,
  title={VideoMultiAgents: A Multi-Agent Framework for Video Question Answering},
  author={Kugo, Noriyuki and Li, Xiang and Li, Zixin and Gupta, Ashish and Khatua, Arpandeep and Jain, Nidhish and Patel, Chaitanya and Kyuragi, Yuta and Ishii, Yasunori and Tanabiki, Masamoto and others},
  journal={arXiv},
  year={2025}
}

@article{liu2026colt,
  title={COLT: Enhancing Video Large Language Models with Continual Tool Usage},
  author={Liu, Yuyang and Cao, Meng and Shi, Xinyuan and Liang, Xiaodan},
  journal={arXiv},
  year={2026}
}

@article{wang2025videochat,
  title={VideoChat-A1: Thinking with Long Videos by Chain-of-Shot Reasoning},
  author={Wang, Zikang and Chen, Boyu and Yue, Zhengrong and Wang, Yi and Qiao, Yu and Wang, Limin and Wang, Yali},
  journal={arXiv},
  year={2025}
}

@article{xu2025vrag,
  title={E-vrag: Enhancing long video understanding with resource-efficient retrieval augmented generation},
  author={Xu, Zeyu and Zhang, Junkang and Wang, Qiang and Liu, Yi},
  journal={arXiv},
  year={2025}
}

@article{xue2025omni,
  title={Omni-AdaVideoRAG: Omni-Contextual Adaptive Retrieval-Augmented for Efficient Long Video Understanding},
  author={Xue, Zhucun and Zhang, Jiangning and Xie, Xurong and Cai, Yuxuan and Liu, Yong and Li, Xiangtai and Tao, Dacheng},
  journal={arXiv},
  year={2025}
}

@article{deepeyes,
  title={DeepEyes: Incentivizing" Thinking with Images" via Reinforcement Learning},
  author={Zheng, Ziwei and Yang, Michael and Hong, Jack and Zhao, Chenxiao and Xu, Guohai and Yang, Le and Shen, Chao and Yu, Xing},
  journal={arXiv},
  year={2025}
}

@inproceedings{hu2025m,
  title={M-llm based video frame selection for efficient video understanding},
  author={Hu, Kai and Gao, Feng and Nie, Xiaohan and Zhou, Peng and Tran, Son and Neiman, Tal and Wang, Lingyun and Shah, Mubarak and Hamid, Raffay and Yin, Bing and others},
  booktitle={CVPR},
  year={2025}
}

@misc{longvt,
      title={LongVT: Incentivizing "Thinking with Long Videos" via Native Tool Calling}, 
      author={Zuhao Yang and Sudong Wang and Kaichen Zhang and Keming Wu and Sicong Leng and Yifan Zhang and Bo Li and Chengwei Qin and Shijian Lu and Xingxuan Li and Lidong Bing},
      year={2025},
      archivePrefix={ArXiv},
}

@misc{molmo,
      title={Molmo2: Open Weights and Data for Vision-Language Models with Video Understanding and Grounding}, 
      author={Christopher Clark and Jieyu Zhang and Zixian Ma and Jae Sung Park and Mohammadreza Salehi and Rohun Tripathi and Sangho Lee and Zhongzheng Ren and Chris Dongjoo Kim and Yinuo Yang and Vincent Shao and Yue Yang and Weikai Huang and Ziqi Gao and Taira Anderson and Jianrui Zhang and Jitesh Jain and George Stoica and Winson Han and Ali Farhadi and Ranjay Krishna},
      year={2026},
      archivePrefix={ArXiv},
}

@inproceedings{nextgqa,
  title={Can i trust your answer? visually grounded video question answering},
  author={Xiao, Junbin and Yao, Angela and Li, Yicong and Chua, Tat-Seng},
  booktitle={CVPR},
  year={2024}
}

@article{zhong2025rethinking,
  title={Rethinking Chain-of-Thought Reasoning for Videos},
  author={Zhong, Yiwu and Hu, Zi-Yuan and Li, Yin and Wang, Liwei},
  journal={Arxiv},
  year={2025}
}

\end{document}